\documentclass{article} %
\usepackage{iclr2023_conference,times}

\usepackage{hyperref}
\usepackage{url}

\usepackage[first=0,last=9]{lcg}
\usepackage{color, colortbl}

\usepackage{amsmath}
\usepackage{listings}
\usepackage{graphicx}
\usepackage{mathtools}
\usepackage{wrapfig}
\usepackage{booktabs}
\usepackage{adjustbox}
\usepackage{subcaption}
\usepackage{xcolor}
\usepackage{amsfonts}
\usepackage{algorithm}
\usepackage{enumitem}
\usepackage{wrapfig}

\definecolor{pearDark}{HTML}{2980B9}
\definecolor{pearDarker}{HTML}{1D2DEC}

\hypersetup{
	colorlinks,
	citecolor=pearDark,
	linkcolor=red,
    breaklinks=true,
	urlcolor=pearDarker}

\newcommand{\method}{\textsc{SemPPL}}

\newcommand{\alpharelic}{\textsc{ReLIC}v2}

\DeclarePairedDelimiterX{\infdivx}[2]{(}{)}{%
  #1\;\delimsize\|\;#2%
}
\newcommand{\kl}{D_{KL}\infdivx}

\DeclareMathOperator*{\mode}{mode}

\definecolor{codegreen}{rgb}{0,0.6,0}
\definecolor{codegray}{rgb}{0.5,0.5,0.5}
\definecolor{codepurple}{rgb}{0.58,0,0.82}
\definecolor{backcolour}{rgb}{0.95,0.95,0.92}

\setcitestyle{square}

\title{SemPPL: Predicting pseudo-labels for better contrastive representations}

\author{Matko Bo\v{s}njak,  Pierre H. Richemond, Nenad Tomasev, Florian Strub, Jacob C. Walker \\ \bf Felix Hill, Lars Holger Buesing, Razvan Pascanu, Charles Blundell, Jovana Mitrovic \\
DeepMind \\
\texttt{\{matko, richemond, mitrovic\}@deepmind.com} \\
}

\iclrfinalcopy %

\lstdefinestyle{mystyle}{
    backgroundcolor=\color{backcolour},   
    commentstyle=\color{codegreen},
    keywordstyle=\color{magenta},
    numberstyle=\tiny\color{codegray},
    stringstyle=\color{codepurple},
    basicstyle=\scriptsize,
    breakatwhitespace=false,         
    breaklines=true,                 
    captionpos=b,                    
    keepspaces=true,                 
    numbers=left,                    
    numbersep=5pt,                  
    showspaces=false,                
    showstringspaces=false,
    showtabs=false,                  
    tabsize=2
}
 
\begin{document}

\maketitle

\begin{abstract}
Learning from large amounts of unsupervised data and a small amount of supervision is an important open problem in computer vision. 
We propose a new semi-supervised learning method, \emph{Semantic Positives via Pseudo-Labels} (\method{}), that combines labelled and unlabelled data to learn informative representations.
Our method extends self-supervised contrastive learning---where representations are shaped by distinguishing whether two samples represent the same underlying datum (positives) or not (negatives)---with a novel approach to selecting positives.
To enrich the set of positives, we leverage the few existing ground-truth labels to predict the missing ones through a $k$-nearest neighbours classifier by using the learned embeddings of the labelled data.
We thus extend the set of positives with datapoints having the same pseudo-label and call these \emph{semantic positives}.
We jointly learn the representation and predict bootstrapped pseudo-labels. This creates a reinforcing cycle. Strong initial representations enable better pseudo-label predictions which then improve the selection of semantic positives and lead to even better representations. 
\method{} outperforms competing semi-supervised methods setting new state-of-the-art performance of $68.5\%$ and $76\%$ top-$1$ accuracy when using a ResNet-$50$ and training on $1\%$ and $10\%$ of labels on ImageNet, respectively. 
Furthermore, when using selective kernels, \method{} significantly outperforms previous state-of-the-art achieving $72.3\%$ and $78.3\%$ top-$1$ accuracy on ImageNet with $1\%$
and $10\%$ labels, respectively, which improves absolute $+7.8\%$ and $+6.2\%$ over previous work.
\method{} also exhibits state-of-the-art performance over larger ResNet models as well as strong robustness, out-of-distribution and transfer performance. We release the checkpoints and the evaluation code at \url{https://github.com/deepmind/semppl}.
\end{abstract}

\section{Introduction}
\label{sec:intro}

In recent years, self-supervised learning has made significant strides in learning useful visual features from large unlabelled datasets \citep{oord2018representation,chen2020simple,mitrovic2020representation,grill2020bootstrap,caron2021emerging}.
Moreover, self-supervised representations have matched the performance of historical supervised baselines on the ImageNet-1k benchmark \citep{russakovsky2015imagenet} in like-for-like comparisons as well as outperformed supervised learning in many transfer settings \citep{relicv2}.
While such results show exciting progress in the field, in 
many real-wold applications often there exists a small amount of ground-truth labelled datapoints making the problem of representation learning semi-supervised.

In this work we propose a novel approach to semi-supervised learning called \emph{Semantic Positives via Pseudo-Labels} (\method{}) which incorporates supervised information during the representation learning stage within a self-supervised loss.
Unlike previous work which uses the available supervision as targets within a cross-entropy objective, we propose to use the supervised information to help inform which points should have similar representations.
We propose to learn representations using a contrastive approach, i.e. we learn the representation of a datapoint (anchor) by maximizing the similarity of the embedding of that datapoint with a set of similar points (positives), while simultaneously minimizing the similarity of that embedding with a set of dissimilar points (negatives).
As such, the appropriate construction of these sets of positives and negatives is crucial to the success of contrastive learning methods.
While strategies for sampling negatives have been extensively studied in the literature \citep{Schroff2015FaceNetAU, Harwood2017SmartMF, Ge2018DeepML, Wang2019MultiSimilarityLW, he2019momentum, Chen2020ImprovedBW}, the sampling of positives has received far less attention.

We propose a novel approach to selecting positives which leverages supervised information.
Specifically, we propose using the small amount of available ground-truth labels in order to non-parametrically predict the missing labels (\emph{pseudo-labels}) for the unlabelled data.
Note that many previous semi-supervised approaches use pseudo-labels as targets within a cross-entropy-based objective \citep{van2020survey, yang2021survey}.
In \method{} we use pseudo-labels in a very different way, i.e. we use them to select positives based on whether two datapoints (we call these \emph{semantic positives}) share the same (pseudo-)label.
By maximizing the similarity of a datapoint with its semantic positives we expect to learn representations that are more semantically aligned and as a consequence encode more abstract, higher-level features which should generalise better.
To predict informative pseudo-labels, we compare the representations of the unlabelled data with those of the labelled subset and use a $k$-nearest neighbours ($k$-NN) classifier to impute the missing labels.

We simultaneously learn the representation, predict pseudo-labels and select semantic positives. 
This creates a \emph{virtuous cycle}: better representations enable better pseudo-label prediction which in turn enables better selection of semantic positives and thus helps us learn better representations. 
Importantly, as the prediction of pseudo-labels and selection of semantic positives does not depend on the exact form of the contrastive objective employed, \method{} is compatible with and complements all contrastive losses, e.g. \citep{chen2020simple,chen2020big,caron2020unsupervised,he2019momentum,mitrovic2020representation} and may even be extended to non-contrastive losses ~\citep{grill2020bootstrap,chen2021exploring}.

We evaluate the representations learned with \method{} across a varied set of tasks and datasets. In particular, \method{} sets new state-of-the-art in semi-supervised learning on ImageNet with $1\%$ and $10$\% of labels on the standard ResNet-50 ($1\times$) architecture with respectively $68.5\%$ and $76.0\%$ top-$1$ performance and across larger architectures.
When combined with Selective Kernels~\citep{Li2019SelectiveKN}, we achieve $72.3\%$ and $78.3\%$ top-1 performance with $1\%$ and $10\%$ labels, respectively, significantly outperforming previous state-of-the-art by absolute $+7.8\%$ and $+6.2\%$ in top-1 performance.
We also outperform previous state-of-the-art on robustness and out-of-distribution (OOD) generalisation benchmarks while retaining competitive performance in transfer learning. 

Our main contributions are:
\begin{itemize}[itemsep=0.05em,topsep=0pt,leftmargin=0.8cm]
    \setlist{nosep}
    \item We extend contrastive learning to the semi-supervised setting by introducing the idea of estimating pseudo-labels for selecting semantic positives as a key component especially in the low-label regime, 
    \item We propose a novel semi-supervised method \method{} that jointly estimates pseudo-labels, selects semantic positives and learns representations which creates a virtuous cycle and enables us to learn more informative representations,
    \item We extensively evaluate \method{} and achieve a new state-of-the-art in semi-supervised learning, robustness and out-of-distribution generalisation, and competitive performance in transfer.
\end{itemize}

\section{Semantic Positives via Pseudo-Labels}
\label{sec:method}

The selection of appropriate positive and negative examples are the cornerstone of contrastive learning. 
Though the research community has mainly focused on the selection of negatives, positives are equally important as they play a vital role in learning semantic similarity. 
We thus leverage labelled information as it encodes semantic information to improve the selection of informative positives.
Specifically, we expand a self-supervised model to use this labelled data to non-parametrically predict pseudo-labels for the remaining unlabelled data. Using both ground-truth labels and the predicted pseudo-labels, we expand the set of positives with semantic positives. 

\textbf{Notations} Let $\mathcal{D}=\mathcal{D}_{l} \cup \mathcal{D}_{u}$ be a dataset consisting of labelled training data $\mathcal{D}_{l} = \{(x_i, y_i)\}_{i=1}^{N}$ and unlabelled training data $\mathcal{D}_{u} = \{(x_j)\}_{j=N+1}^{M}$ with $M \gg N$. Let $\mathcal{B}$ be a batch of data of size $B$ with $\mathcal{B} = \{ (x_i, y_i) \}_{i=1}^{b} \cup \{ x_j \}_{j=b+1}^{B}$ where $(x_i, y_i) \in \mathcal{D}_{l}$ and $x_{j} \in \mathcal{D}_u$, where the indices $i$, $j$ and $m$ to denote labelled, unlabelled, and all datapoints, respectively. Following established self-supervised learning practices \citep{chen2020simple,chen2020big,caron2020unsupervised,mitrovic2020representation,dwibedi2021little, relicv2}, we create different views of the data by applying pairs of randomly sampled augmentations $a_1, a_2 \sim \mathcal{A}$ from the augmentation distribution $
\mathcal{A}$ proposed in \citet{chen2020simple}. 
For every datapoint $x_{m} \in \mathcal{D}$ we denote the corresponding augmentations as $x_{m}^{a_1}, x_{m}^{a_2}$. 

\paragraph{Augmentation positives}
We embed one data view $x_{m}^{a_1}$ via an \emph{online} encoder network $f$ and embed the other data view $x_{m}^{a_2}$ with a \emph{target} encoder network $f_{t}$, i.e. we get latent representations $z_{m}^{a_1}=f(x_{m}^{a_1})$ and $z_{m, t}^{a_2}=f_{t}(x_{m}^{a_2})$.
Note that the weights of $f_{t}$ are an exponential moving average of the weights of $f$. 
Next, we pass these latent representations through projection and prediction multi-layer perceptrons.
Specifically, we use an online projector $g$ and target projector $g_{t}$, as well as an online predictor $h$, to further transform $z_{m}^{a_1}$ and $z_{m, t}^{a_2}$; again, the weights of $g_{t}$ are an exponential moving average of the weights of $g$.
We then get $\hat{z}_m^{a_1} = h(g(z_m^{a_1}))$ and $\tilde{z}_{m, t}^{a_2} = g_{t}(z_{m, t}^{a_2})$ and $l_2$-normalise these; we use $\hat{z}_m^{a_1}, \tilde{z}_{m, t}^{a_2}$ onward as the normalised latent embeddings.

In order to learn the representation of $\hat{z}_m^{a_1}$, we contrast it against the augmentation-based positive $\tilde{z}_{m, t}^{a_2}$ as well as against negatives. For this, we use the contrastive loss:
\begin{alignat}{1}\label{loss_augpos}
\mathcal{L}_{\text{\textsc{augm}}} = - \sum_{m=1}^{B} 
        \log 
        \frac{\varphi(\hat{z}_m^{a_1}, \tilde{z}_{m, t}^{a_2})}{\varphi(\hat{z}_m^{a_1}, \tilde{z}_{m, t}^{a_2}) + 
        \sum_{x_n \in\mathcal{N}(x_m)}\varphi(\hat{z}_m^{a_1},\tilde{z}_{n, t}^{a_2})}
\end{alignat}
where $\mathcal{N}(x_{k})$ is the set of negatives, randomly uniformly sampled from the current batch, $\tilde{z}_{n, t}^{a_2} = g_{t}(f_{t}(x_{n}))$  the target network projection of the negative sample; $\varphi(x_1, x_2) = \tau \cdot \exp(\langle x_1, x_2 \rangle / \tau)$ is the scoring function, $\tau>0$ is a scalar temperature, and $\langle \cdot, \cdot \rangle$ denotes the Euclidean dot product. Since the representations we contrast are $l_2$-normalised, the dot product effectively turns into cosine similarity.

\begin{figure}[t]
    \centering
    \includegraphics[width=1\linewidth]{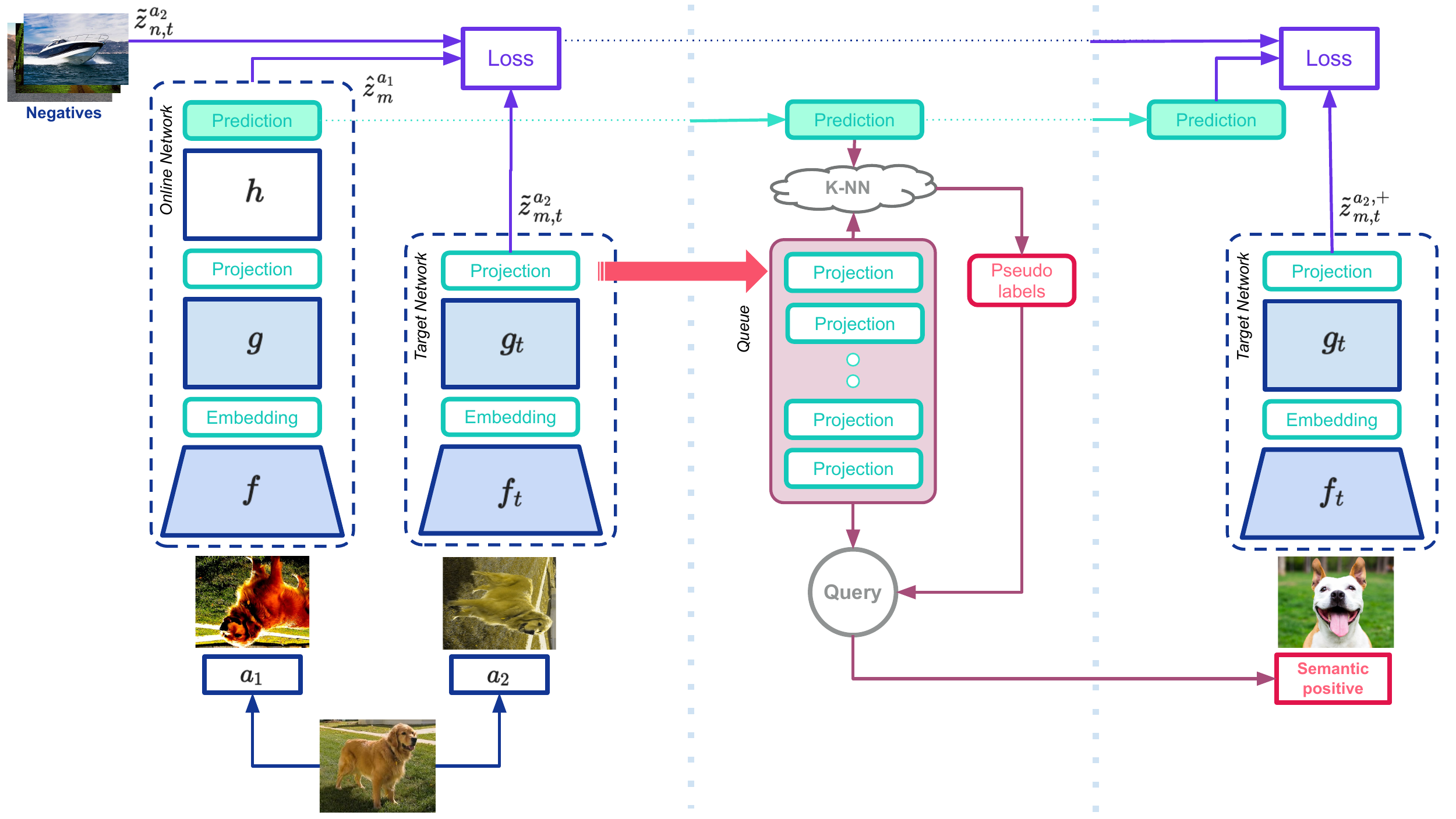}
    \caption{Sketch of \method{}. (Left) Standard contrastive pipelines. (Middle) Unlabelled data are tagged with pseudo-labels by using a $k$-NN over projected labelled data. (Right) Semantic positives are queried from the queue and processed to compute an additional contrastive loss.}
    \label{fig:model}
\end{figure}

\paragraph{Pseudo-label prediction and semantic positives} 
Since we have access to a small labelled dataset, we can use the label information to select more informative positives beyond just augmentations of the original image. 
Specifically, we can associate images with the same label as positives and we call these \emph{semantic positives}.
We want to select semantic positives for all the data, not just the labelled subset.
For this purpose, we propose to compute pseudo-labels for the unlabelled data and use this to select semantic positives.
To compute pseudo-labels we compare the current latent embeddings of the unlabelled data to those of the labelled data. 
Specifically, we propose to use a first-in-first-out queue $Q$ with capacity $C$ for storing labelled embeddings which we use for computing the pseudo-labels.
At the start of training, we simply initialise the queue with random vectors, and use the queue from the first step.
For each batch $\mathcal{B}$, we add the target projection of only the labelled data to the queue, i.e.\ $Q \leftarrow (\tilde{z}_{i, t}^{a_2}, y_i)$.
To predict a pseudo-label for an unlabelled datapoint $x_j$, we first compute the online predictor output $\hat{z}_j^{a_1}$, before retrieving its $k$-nearest neighbours $\{ (\tilde{z}_{s, t}^{a_2}, y_s) \}_{s=1}^{k}$ in cosine similarity from the queue $Q$.\footnote{We use the cosine similarity as the embeddings are normalised.}
Finally, we compute the pseudo-label $\bar{y}_{j}$ of $x_j$ as:
\begin{equation}
    \bar{y}_{j} = \mode_{y_s} \{ (\tilde{z}_{s, t}^{a_2}, y_s) \}_{s=1}^{k}
\end{equation}
where $\mode$ is the mode of the set, tasked with obtaining the most frequent class in the k-nearest neighbours.
We use the ground-truth labels (for the labelled data) or the computed pseudo-labels (for the unlabelled data) to select \emph{semantic positives} for every datapoint in $\mathcal{B}$.
For each $x_m \in \mathcal{B}$, we uniformly sample over all the embeddings in $Q$ that share the same (pseudo-) label as $x_m$ to get a semantic positive $\tilde{z}_{m, t}^{a_2, +} \sim U(\{ (\tilde{z}_{l, t}^{a_2}, y_l) \in Q \ |\ y_l = pl(x_m) \})$, where $pl(x_m) = y_m$ if $x_m$ is labelled and $pl(x_m) = \bar{y}_m$ if $x_m$ is unlabelled.
Next, we include these semantic positives within our representation learning process through the contrastive objective
\begin{alignat}{1}\label{loss_sempos}
\mathcal{L}_{\text{\textsc{sempos}}} = - \sum_{m=1}^{B} 
        \log 
        \frac{\varphi(\hat{z}_m^{a_1}, \tilde{z}_{m, t}^{a_2, +})}{\varphi(\hat{z}_m^{a_1}, \tilde{z}_{m, t}^{a_2, +}) + 
        \sum_{x_n \in\mathcal{N}(x_m)} \varphi(\hat{z}_m^{a_1}, \tilde{z}_{n, t}^{a_2})}
\end{alignat}
Taking these two losses (\ref{loss_augpos}) and  (\ref{loss_sempos}) together, we propose to learn representations in our method SemPPL by minimising the following total loss
\begin{equation}\label{eq:final_loss_sempos}
    \mathcal{L}_{\text{\method}} = \mathcal{L}_{\text{\textsc{augm}}}  + \alpha\mathcal{L}_{\text{\textsc{sempos}}}
\end{equation}
where $\alpha$ controls the ratio between these sub-losses.

\subsection{Implementation details}

\textbf{Architecture} We use Residual Networks  \citep{he2016deep} (v1; pre-activation as customary in the literature) for $f$ and $f_{t}$ and use either $50$ or $200$ layers deep networks and with a width multiplier ranging from $1\times$ to $4\times$. As in~\citep{grill2020bootstrap, relicv2}, we use multi-layer perceptrons with $2$ layers of size $4096$ and $256$, with batch normalisation~\citep{Ioffe2015BatchNA} and rectified linear activation.

\textbf{Self-supervised learning method}
We use \alpharelic{}~\citep{relicv2} as our default self-supervised training objective due to its competitive performance. 
Therefore, we add an invariance penalty on top of Equation~\ref{eq:final_loss_sempos} to further enforce the similarity constraints and regularize the learning process as detailed in Appendix~\ref{app:sec:invariance}. We also explore other self-supervised learning objectives in Section~\ref{sec:analysis}.

\textbf{Algorithm parameters} We use a queue of capacity $C=20 B$, with batch size $B=4096$, and temperature $\tau=0.2$ while randomly sampling negatives from the current batch; we take $|\mathcal{N}(x)| = 10$ negatives in total. 
For augmentations, we use the standard \textsc{SimCLR} augmentations~\citep{chen2020simple} and the \textsc{ReLICv2} multi-crop and saliency-based masking~\citep{relicv2}; we use $4$ large views and $2$ small views for augmentation positives and $3$ semantic positives.
The semantic positives are computed with a $k$-NN with $k=1$ (see the analysis section in Appendix~\ref{analysis_appendix}); we build a single $k$-NN instance per augmentation $a$ queried with all the augmentations where $|a|=4$. 
This produces $|a|^2=16$ $k$-NN induced pseudo-labels in total for each unlabelled image among which we then perform majority voting to compute the final pseudo-label.

\textbf{Optimisation} Our networks are optimized with \textsc{LARS}~\citep{you2017large}. Our base learning rate is $0.3$ and we train our models for $300$ epochs with a learning rate warm-up period of $10$ epochs and cosine decay schedule thereafter.
We use a weight decay of $10^{-6}$ and batch size $B = 4096$. We exclude the biases and batch normalisation parameters both from \textsc{LARS} adaptation and weight decay. The exponential moving average parameter for target networks is $0.996$. Our pseudo-code is described in the appendix along with precise architectural and implementation details. Pretrained model checkpoints and code are available at \url{https://github.com/deepmind/semppl}.

\section{Experimental Results}
\label{sec:results}

To evaluate \method{}, we pre-train representations using 1\% and 10\% labelled data from the ImageNet dataset~\citep{russakovsky2015imagenet} based on the splits from \citet{chen2020simple}.
We then test \method{} in semi-supervised classification, robustness and out-of-distribution generalisation tasks.
Lastly, we probe the transfer capabilities of the representations to other image classification datasets.
For a complete set of results and experimental details, please see the Appendix~\ref{results_app}.

\subsection{Semi-supervised learning}
\label{sec:semi-supervised-learning}

In Table \ref{table.semi-sup}, we report top-1 accuracy on the ImageNet test set when either 1\% or 10\% of the data is labelled for the ResNet-$50$ architecture as well as deeper and wider ResNets. 
\method{} achieves top-$1$ accuracy of $68.5\%$ with 1\% of labels, significantly outperforming the previous state-of-the-art SimMatch~\citep{Zheng2022SimMatchSL} by an absolute $+1.3\%$ in ImageNet test accuracy. With 10\% of label data, our top-1 accuracy on ResNet-$50$ reaches $76.0\%$, outperforming the previous state-of-the-art PAWS~\citep{assran2021semi} in semi-supervised learning.
\method{} outperforms competing representation learning methods across the board, achieving state-of-the-art performance on all ResNet-50 $2\times$, ResNet-50 $4\times$ and , in both the $1\%$ and $10\%$ labelled settings. \method{} does not use, and therefore excludes from comparison, distillation from larger networks as in~\citep{chen2020big,pham2021meta}.

Similar to~\citep{chen2020big}, we also tested \method{} on ResNets with \emph{Selective Lernels} (SK)~\citep{Li2019SelectiveKN}.
This increases the encoder parameter count to $27.5$M.
We thus achieve a new absolute state-of-the-art of $72.3\%$ and $78.3\%$ top-$1$ accuracies, respectively, when using $1\%$ and $10\%$ of labelled data. 
Finally, \method{} reaches a new state-of-the-art using 76.0 and 80.5 on 1\% and 10\% of labels without self-distillation with a ResNet-200 $2\times$ + SK architecture. 

For implementation details of the semi-supervised results and additional results, see the Appendix~\ref{app.semi_sup_results}.

\begin{table}[ht]
    \centering
    \caption{Top-$1$ accuracy (in \%) for ResNet encoders with different depth and width.}\label{table.semi-sup}
    \small
\begin{adjustbox}{max width=\columnwidth}
\begin{tabular}{lccccccccc}
\toprule
& \multicolumn{2}{c}{ResNet-50 $1 \times $} & \multicolumn{2}{c}{ResNet-50 $2 \times $} & \multicolumn{2}{c}{ResNet-50 $4 \times $} & \multicolumn{2}{c}{ResNet-200 $2 \times $} \\
\midrule
Method  &  \multicolumn{2}{c}{Top-1} &  \multicolumn{2}{c}{Top-1} &  \multicolumn{2}{c}{Top-1} &  \multicolumn{2}{c}{Top-1} \\ %
 &  1\% & 10\% &  1\% & 10\% & 1\% & 10\% & 1\% & 10\% \\ %
\midrule
    \quad SimCLR \citep{chen2020simple}     & 48.3 & 65.6   & 58.5 & 71.7 & 63.0 & 74.4 & - & -\\ %
    \quad BYOL \citep{grill2020bootstrap}   & 53.2 & 68.8   &  62.2 & 73.5 & 69.1 & 75.7 & 71.2 & 77.7 \\ %
    \quad \alpharelic{} \citep{relicv2}     & 58.1 & 72.4   & 64.7 & 73.7 & 69.5 & 74.6 & 72.1 & 76.4 \\ %
    \quad SimCLRv2 \citep{chen2020big}      & 57.9 & 68.4   & 66.3 & 73.9 & - & - & - & - \\
    \quad CoMatch~\citep{Li2021CoMatchSL}   & 66.0 & 73.7   & - & - & - & - & - & - \\
    \quad PAWS \citep{assran2021semi}       & 66.5 & 75.5   & 69.6 & 77.8 & 69.9 & 79.0 & - & - \\
    \quad SimMatch~\citep{Zheng2022SimMatchSL} & 67.2 & 74.4 & - & - & - & - & - & - \\
    \rowcolor{blue!10} \quad SemPPL (ours)                  & {\bf 68.5} & {\bf 76.0}   & {\bf 71.9} & {\bf 78.6} & {\bf 72.5} & {\bf 79.3} & {\bf 74.8} & {\bf 80.4} \\ 
    \midrule
    \quad SimCLRv2 + SK~\citep{chen2020big} & 64.5 & 72.1 & 70.6 & 77.0 & - & - & - & -  \\
    \rowcolor{blue!10} \quad SemPPL + SK (ours) & {\bf 72.3} & {\bf 78.3} & {\bf 74.5} & {\bf 79.8} & - & - & {\bf 76.0} & {\bf 80.5}  \\
\bottomrule
\end{tabular}
\end{adjustbox}
\end{table}

\subsection{Robustness and OOD generalisation}
\label{sec:res:robust}

We evaluate the robustness and generalisation abilities of \method{} on ImageNetV2 \citep{recht2019imagenet}, ImageNet-C \citep{hendrycks2019benchmarking}, ImageNet-R \citep{hendrycks2021many} and ObjectNet \citep{barbu2019objectnet} which have all been purposefully constructed to test different robustness and generalisation aspects. 
We evaluate all three variants on ImageNetV2: matched frequency (MF), Threshold 0.7 (T-0.7) and Top Images (TI).
When evaluating PAWS, we used the publicly available checkpoints.
Table~\ref{tab.ood} shows good robustness and generalisation ability of the representations learned with \method{}.
\method{} sets the new state-of-the-art performance (outperforming even the supervised baseline) on $4$ out of $5$ datasets, while outperforming PAWS across all datasets. \method{} also outperforms SimMatch on $4$ out of $5$ datasets.
For more details on the evaluation protocols and results for ImageNet-C see the Appendix~\ref{app.robust_ood}.

\begin{table}[ht]
\small 
  \centering
    \caption{Top-1 accuracy (in \%) for ImageNetV2, ImageNet-R and ObjectNet.}
      \label{tab.ood}
  \small
  \begin{tabular}{lccccc}
  \toprule
  & \multicolumn{3}{c}{Robustness} & \multicolumn{2}{c}{OOD generalization} \\
  \midrule
  Method & MF & T-0.7 & Ti & ImageNet-R & ObjectNet  \\
  \toprule
  \quad Supervised (100\% labels) ~\tiny\citep{lim2019fast}  & 65.1  & 73.9 & 78.4  & 24.0  & {\bf 26.6}  \\ %
  \midrule
    \emph{Semi-supervised (10\% labels)} \\
    \quad PAWS~\citep{assran2021semi} & 64.5 & 73.7 & 78.9 &  23.5 & 23.8 \\
    \quad SimMatch~\citep{Zheng2022SimMatchSL} & 63.8 & 73.2 & 78.3 &  {\bf 25.0} & 24.5 \\ 
    \rowcolor{blue!10} \quad SemPPL (ours)  & {\bf 65.4} & {\bf 74.1} & {\bf 79.6} & 24.4 & {\bf 25.3} \\
  \bottomrule
  \end{tabular}
\end{table}

\subsection{Transfer learning}

We evaluate the generality of \method{} representations by testing whether the features learned on ImageNet are useful across different datasets and tasks.
Specifically, we evaluate the transfer performance of \method{} on a set of $11$ image classification datasets commonly used in the contrastive literature under the linear protocol \citep{grill2020bootstrap,chen2020simple,dwibedi2021little,mitrovic2020representation,relicv2}.
For the linear protocol, the pretrained encoder is frozen and a  randomly initialized linear classifier is trained on top using the training data from the target dataset.
We report standard metrics for each dataset as well as performance on a held-out test set. 
For more details on the evaluation protocol see the Appendix~\ref{sec:supp:transfer}.
Table~\ref{table.transfer} compares the transfer performance of representations pretrained using the supervised baseline~\citep{chen2020simple}, PAWS~\citep{assran2021semi}, SimMatch~\citep{Zheng2022SimMatchSL} and our method \method{}.
\method{} outperforms the supervised baseline on $8$ out of $11$ datasets, PAWS on $9$ out of $11$ datasets, while showing competitive performance to SimMatch, outperforming it on $4$ out of $7$ datasets.

\begin{table}[t!]
\caption{Top-1 accuracy (in \%) on the full suite of transfer tasks.}
\label{table.transfer}
\begin{center}
\setlength\tabcolsep{4pt}
\begin{adjustbox}{max width=\columnwidth}
\begin{tabular}{lccccccccccc}
\toprule
Method & Food101 & CIFAR10 & CIFAR100 & Birdsnap & SUN397  & Cars & Aircraft & DTD & Pets & Caltech101 & Flowers \\
\toprule
\; Supervised-IN \citep{chen2020simple} & 72.3 & {\bf 93.6} & 78.3 & 53.7 & 61.9 & 66.7 &  61.0 & 74.9 & 91.5  &  {\bf 94.5} & 94.7 \\
\midrule
\emph{Semi-supervised (10\% labels)}\\
\; PAWS~\citep{assran2021semi} & 79.1  &  92.3 & 76.3 & 62.0  &  66.1 & {\bf 75.7}  &  61.4  &  77.0 & 92.2  &  91.9  & {\bf 96.5} \\
\; SimMatch~\citep{Zheng2022SimMatchSL} & 71.7 & {\bf 93.6} & {\bf 78.4} & -- & -- & 69.7 & -- & 75.1 & {\bf 92.8} & -- & 93.2 \\
\rowcolor{blue!10} \; \method{} (ours) & \textbf{80.2} & 92.5 & 77.6 & \textbf{64.2} & {\bf 66.3} & 75.5 & \textbf{63.9} & \textbf{77.8} & 92.5 & 93.0 & 96.3 \\
\bottomrule
\end{tabular}
\end{adjustbox}
\end{center}
\end{table}

\subsection{Full labelled dataset}

\begin{wrapfigure}{r}{0.6\textwidth}
    \small
    \caption{Top-$1$ accuracy for ResNet50 with 100\% of the labels across augmentations, initializations and networks.}
    \label{table.full_labels}
    \begin{tabular}{l c c}
    \toprule
    Method & Params & Top-1 \\
    \midrule 
    \emph{Supervised (ResNet-50)} & & \\
    \quad + AutoAugment~\citep{cubuk2018autoaugment} & 27M & 77.6 \\
    \quad + MaxUp~\citep{gong2020maxup}             & 27M & 78.9 \\
    \midrule
    \emph{Representation Learning (ResNet-50)} & & \\
    \quad \method{} (SimCLR base) & 27M & 76.0 \\
    \quad \method{} (BYOL base)   & 27M & 77.7 \\
    \rowcolor{blue!10} \quad \method{} (ReLICv2 base; ours) & 27M &  79.7 \\
    \midrule
    \emph{Other Architectures} & & \\
    \quad Swin-T~\citep{liu2021swin}       & 29M & 81.3 \\
    \quad ConvNeXt~\citep{liu2022convnet}  & 29M & 82.1 \\
    \rowcolor{blue!10} \quad \method{} + SK (ours) & 29M & 82.0 \\
    \bottomrule
    \end{tabular}
\end{wrapfigure}

We also assess how \method{} behaves in a fully supervised setting.
For this purpose, we select semantic positives based on the ground-truth labels and fine-tune the learned representations with the full ImageNet dataset.
We compare against strong supervised baselines on ResNets as well as against recent performant network architectures that are extensions of the ResNet, e.g. \citep{liu2021swin, liu2022convnet}.
Our method reaches $79.7\%$ top-1 accuracy on a ResNet 50 outperforming a number of strong supervised baselines. When we add selective kernels to a ResNet 50, we achieve $82\%$ top-1 accuracy outperforming recent transformers architecture~\citep{liu2021swin}, and matching highly tuned ConvNext~\citep{liu2022convnet}. Therefore, \method{} may also be considered as a promising pretraining method in the supervised learning setting.

\section{Analysis}
\label{sec:analysis}

We analyse the impact of different design choices in \method{} on downstream performance. 
In this section, we focus the behaviour and impact of pseudo-labels and semantic positives on learning representations.
For further analyses and experimental details, please see Appendix~\ref{analysis_appendix}.

\paragraph{Semantic positives across self-supervised learning objectives}
With \method{} we extend the set of positives to include semantic positives based on predicted pseudo-labels; we can combine these ideas with other self-supervised methods. In Table~\ref{tab.ablation.1}, we additionally evaluate \method{} on the non-contrastive self-supervised method BYOL~\citep{grill2020bootstrap}.
BYOL replaces the contrastive loss in Equation~\ref{eq:final_loss_sempos} with an $l_2$ loss.
Importantly, we follow the training pipeline (e.g. augmentation, hyperparameters etc.) from \citep{grill2020bootstrap} to fairly highlight the impact of \method{}. 
We observe a drastic improvement when adding semantic positives.
With 1\% labels on ImageNet BYOL improves by absolute $+3.9\%$ and by absolute $+3.6\%$ when using $10\%$ labels. 
For completeness we have also highlighted the contribution of \method{} when using \alpharelic{} as the base self-supervised objective which is our default implementation.
For $1\%$ labeled data, we see an absolute improvement of $+10.4\%$ in top-1 accuracy, while for $10\%$ labels we see a gain of absolute $+3.6\%$ in top-1 accuracy.
In summary, we see that \method{} can be easily combined with other self-supervised objectives to yield significant improvements and can be used a plug-and-play module in semi-supervised learning.

\paragraph{The contribution of pseudo-labels and semantic positives} We examine the impact of omitting pseudo-label prediction and semantic positives from learning representations.
Specifically, we ablate the use of pseudo-labels when selecting semantic positives for unlabelled datapoints, i.e. we only use labelled images when retrieving semantic positives.
In Table~\ref{tab.ablation.2} (middle row), removing pseudo-label prediction significantly decreases performance both in the $1\%$ and $10\%$ label settings. In addition, the low-label regime ($1\%$ labels) suffers a stronger performance decrease $-6.6\%$ than the $10\%$ labels regime, $-4.9\%$.
This underscores the importance of pseudo-label estimation and subsequent selection of semantic positives for unlabelled data especially in the low-data regime.
Going a step further, we remove semantic positives even for the labelled data, falling back to vanilla \alpharelic{}. In Table~\ref{tab.ablation.2} (bottom row), we see again a significant drop in performance for both the $1\%$ and $10\%$ label settings with a sharper drop for the low-label regime.
Together these highlights the importance of both including semantic positives for labelled data as well as using pseudo-label prediction for selecting semantic positives for unlabelled data in order to learn informative representations in a semi-supervised setting in a label-efficient way.

\begin{table}[b!]
    \caption{Top-1 test accuracy (in \%) with a ResNet50 pretrained on ImageNet with 1\% and 10\% labels.}
    \label{tab.ablation}
    \begin{subtable}{.48\linewidth}
        \small
        \begin{center}
        \begin{tabular}{ l  l  l}
        \toprule
                                            & \multicolumn{2}{c}{Top-1} \\ 
                                            &   1\%  &  10\%    \\
        \midrule
         BYOL~\citep{grill2020bootstrap}        &  53.2  &  68.8  \\
         \method{} with BYOL                       &  57.1  &  72.4 \\ %
        \midrule
         ReLICv2~\citep{relicv2}        &  58.1  &  72.4  \\
        \rowcolor{blue!10}  \method{} with \alpharelic{} (ours)     &  \textbf{68.5}  &  \textbf{76.0} \\  %
        \bottomrule
        \end{tabular}
        \subcaption{Trained on a different self-supervised objective.}
        \label{tab.ablation.1}
    \end{center}
    \end{subtable}
    \hfill
    \begin{subtable}{.48\linewidth}
        \setlength\tabcolsep{5pt}
        \small
        \begin{center}
        \begin{tabular}{ l  c  c}
        \toprule
                                         &   1\% labels  &    10\% labels  \\
        \toprule
        \rowcolor{blue!10}   \method{}                       &   {\bf 68.5} & {\bf 76.0} \\
         \quad - Pseudo-labels           &   61.9 & 71.1 \\  
         \quad - Semantic Positives      &   58.1 & 72.4 \\   
        \bottomrule
        \end{tabular}
        \subcaption{Removing pseudo-labelling and semantic positives in \method{}.}
        \label{tab.ablation.2}
        \end{center}
    \end{subtable}
\end{table}

\begin{figure}[ht]
    \centering
    \includegraphics[width=0.95\linewidth]{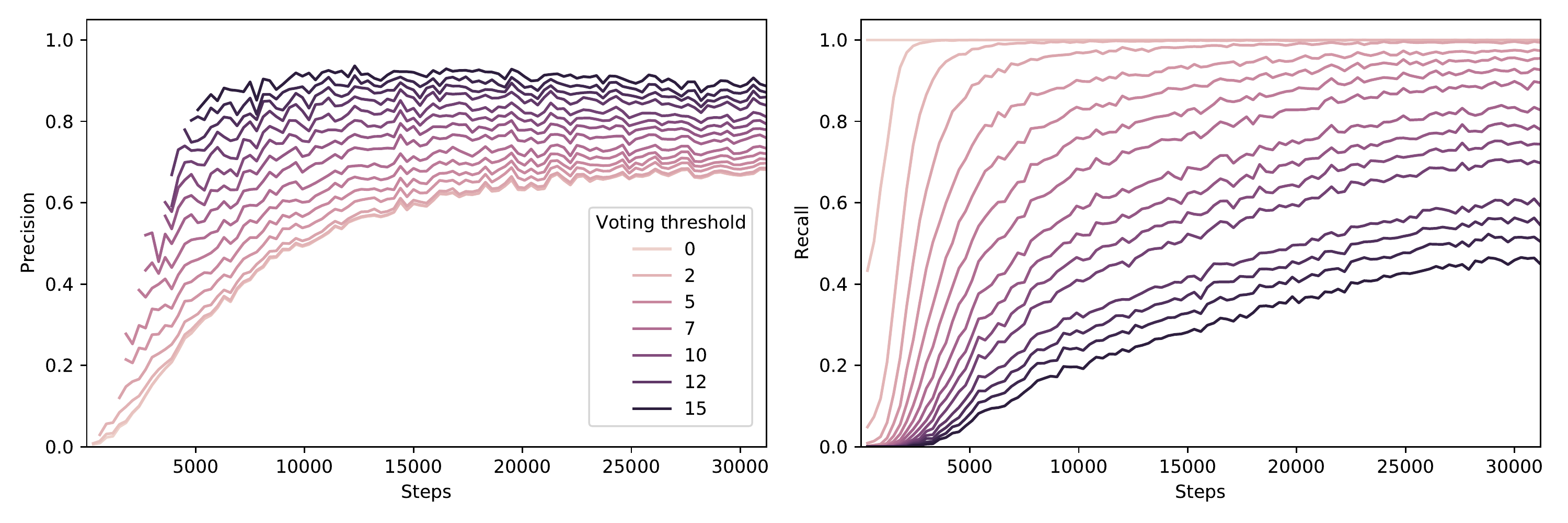}
    \caption{Precision and recall for pseudo-labels computed based on $k$-nearest neighbours when trained on ImageNet with 10\% labels over 100 epoches.}
    \label{fig:precision-recall}
\end{figure}

\paragraph{Precision and Recall of pseudo-labels.} In Figure~\ref{fig:precision-recall}, we analyse the behaviour of pseudo-labels  by looking at the precision and recall as training progresses.
We train a ResNet-50 for $100$ epochs using $10\%$ labels with \method{} on ImageNet.
As we have $4$ large views there will be in total $16$ votes cast and then the pseudo-label will be estimated using majority voting.
We want to measure how often these $16$ votes agree or disagree; we denote as voting threshold the number $k$ where at least $k$ votes have been cast for one class.
We see that as training progresses the precision across all thresholds increases as expected.
This means that the pseudo-label prediction is bootstrapping itself to become more accurate, which enables us to select better semantic positives and thus learn more informative representations as training progresses, i.e. we have a virtuous cycle of representation learning and pseudo-label prediction.
Furthermore, precision is an increasing function of the voting threshold throughout training and is highest for the biggest voting threshold.
This indicates how confident we can be in the accuracy of pseudo-label prediction, and thus how confident we can be that an appropriate semantic positive has been selected.
Yet, we see that the recall for individual thresholds is also increasing as training progresses but that the recall decreases as we increase the voting threshold. 
This is expected as there is always a trade-off between precision and recall.

\paragraph{Noise in pseudo-label prediction}
In Figure~\ref{fig:precision-recall}, we observe that the proportion of correctly predicted pseudo-labels at the end of training is reasonably high ($60\%$ accuracy of voting threshold $0$). Yet, it also means that $40\%$ of pseudo labels are still incorrectly predicted. As the incorrect prediction results in suboptimal semantic positives selection, i.e., \method{} does not select semantic positives from the same class as the datapoint, this behavior may ultimately worsen the quality of extracted features.
To quantify this phenomenon, we train the representation with \method{} where an oracle replaces the pseudo-label prediction with ground-truth labels and those are used for selecting semantic positives. %
In Table~\ref{tab.oracle}, we train the representations for $100$ epochs on ImageNet with $10\%$ labels.
There, the oracle increases the top-1 performance of 1.7\% in the test set with 10\%. Besides, the pseudo-label accuracy also gets 6.2\% higher. It thus confirms that incorrect pseudo-label predictions, and incorrect semantic positives retrieval, hurts learning informative representations and the downstream performance. 
Yet, the oracle performance remains close to the actual performance of \method{}, illustrating the method's robustness.

\begin{table}
    \caption{ Top-1 test accuracy (in \%) with a ResNet50 pretrained on ImageNet 10\% labels for 100 epoches when using ground truth labels instead of pseudo-labels while retrieving semantic positives (oracle); PL accuracy holds for pseudo-label accuracy.}
    \label{tab.oracle}
    \centering
        \begin{tabular}{ l  c  c}
        \toprule
        10\% labels                                 &   Top-1  &   PL accuracy   \\
        \toprule
        \rowcolor{blue!10}   \method{}                       &   69.9  &  69.7 \\
         \method{} (+oracle)             & \textbf{71.6} &  \textbf{76.9} \\
        \bottomrule
        \end{tabular}
\end{table}

Further ablations on other design choices, such as the number of semantic positives, the use of view voting, the choice of $k$ in $k$-NN, queue length and training duration can be found in Appendix~\ref{additional_analysis_appendix}.

\section{Related work}
\label{sec:review}

\paragraph{Semi-supervised learning} In the semi-supervised regime~\citep{cheplygina2019not, van2020survey, yang2021survey, alizadehsani2021uncertainty}, one can either pre-train a model on unlabelled data and subsequently fine-tune it on labelled data, or train both jointly. Joint training on labelled and unlabelled data often involves combining the two losses~\citep{grandvalet2006entropy, miyato2018virtual, zhai2019s4l, verma2019interpolation, berman2019multigrain, xie2020unsupervised}. Pseudo-label self-training approaches~\citep{zoph2020rethinking} present an important alternative, first inferring approximate pseudo-labels for the unlabelled examples, and then incorporating them in supervised losses. Pseudo-labels can either be generated prior to a subsequent supervised learning phase~\citep{yarowsky1995unsupervised, riloff1996automatically, lee2013pseudo} or jointly in an online fashion~\citep{berthelot2019mixmatch, berthelot2019remixmatch, sohn2020fixmatch}.
These methods may benefit from pseudo-label confidence measures~\citep{sohn2020fixmatch, pseudodefence, zhang2021flexmatch} as well as thresholding~\citep{xu2021dash}, temporal ensembling~\citep{Laine2017TemporalEF}, or stronger regularization to mitigate bias in early model training~\citep{sajjadi2016regularization, arazo2020pseudo}. The use of pseudo-labels with rebalancing has shown improvements, both in class-imbalanced problems~\citep{Wei_2021_CVPR} and in a general context~\citep{Wang2022DebiasedLF}. Teacher-student network configurations for generating and utilising pseudo-labels have also shown promise~\citep{tarvainen2017mean, Luo2018SmoothNO, Ke_2019_ICCV, xie2020self, Cai_2021_CVPR, pham2021meta}. Co-training  uses different feature extractors for different data views and alternates between pseudo-labelling and training phases~\citep{blum1998combining, Qiao2018DeepCF}.
Good performance has been reached by using consistency losses between pseudo-labels of different inputs~\citep{verma2019interpolation, Hu_2021_CVPR}. 

Predicting view assignments with support samples~\citep{assran2021semi} (PAWS) has resulted in substantial performance improvements, with the idea that the assigned pseudo-labels ought to be similar across multiple views of the same image. 
Recent work has shown that incorporating label information in positive selection in contrastive methods is highly promising, compared to the cross-entropy loss in the fully supervised case~\citep{khosla2020supervised}. Our method demonstrates a similar utility of pseudo-labels for semi-supervised problems, and differs from competing ones in the following ways. Unlike DebiasPL \citep{Wang2022DebiasedLF} that uses an adaptive margin loss, SemPPL does not seek to directly address or re-shape the distribution of pseudo-labels. Unlike SimCLRv2 \citep{chen2020big}, we do not rely on self-distillation procedures. In contrast with PAWS \citep{assran2021semi}, we fully leverage the contrastive approach for semi-supervised learning; not using positives only for training means \method{} does not require specific care like pseudo-labels sharpening to stabilize learning and avoid representational collapse. \method{} is more closely related to CoMatch \citep{Li2021CoMatchSL} that also uses bootstrapping to improve pseudo-labels representational quality, but is conceptually much simpler, avoiding phases of distributional alignment and of performing graph-based contrastive learning. In a similar vein, SimMatch \citep{Zheng2022SimMatchSL} also uses a memory buffer to propagate pseudo-labels, but has a more complex objective than \method{} and equally requires additional phases of pseudo-labels unfolding and aggregation to function.

\paragraph{Self-supervised learning} Major advances in learning useful representations from unlabelled data~\citep{liu2021self, ssl_wild} can be seen as a paradigm shift, since these methods have recently been competitive with supervised training baselines~\citep{relicv2}. A number of self-supervised learning methods involve contrasting multiple views of the data~\citep{oord2018representation, bachman2019learning, chen2020simple, he2019momentum, grill2020bootstrap, dwibedi2021little}.
Similar performance were also achieved by bootstrapping-based multi-view learning \citep{grill2020bootstrap, richemond2020byol, chen2021exploring, Zbontar2021BarlowTS, Wang2021TowardsDR}, or involving explicit clustering steps~\citep{caron2020unsupervised, Asano2020SelflabellingVS, Li2021PrototypicalCL}. 
An explicit causally-motivated invariance loss, when used in conjunction with the contrastive objective, has been shown to lead to more compact representations, and desirable generalisation properties~\citep{mitrovic2020representation, relicv2}. 
Contrastive approaches are not always used in self-supervised methods~\citep{he2021masked, ermolov2021whitening, chen2022context}. Transformer-specific methods have been devised~\citep{caron2021emerging, chen2021empirical,zhai2022position}.

\section{Conclusion}
\label{sec:conclusion}
In this work, we propose \method{}, a novel semi-supervised learning method to incorporate semantic positives in self-supervised objectives by taking advantage of pseudo-labels.
Through extensive empirical evaluation, we demonstrated that our approach achieves state-of-the-art semi-supervised performance on ImageNet across several ResNet architectures as well as on the robustness, out-of-distribution generalization and transfer tasks.
We also show that \method{} can be easily combined with other existing self-supervised methods and is a promising direction to pre-train networks also in a fully supervised learning regime.
Our analyses suggest that the role of pseudo-labels in selecting positives for semi-supervised contrastive methods might be underappreciated.  Despite widespread use in semi-supervised applications, pseudo-labels are less understood and have been explored far less in the context of self-supervised methods.  We hope this study, which shows empirically that prior work has under-utilized pseudo-labels, may help bridge that gap. 

\newpage
\section*{Reproducibility Statement}

We documented the model and experimental evaluation in the main body of the paper and added further details in the appendix.
Concretely, we explain implementation details and sweep parameters in Appendix~\ref{results_app} and Appendix~\ref{additional_analysis_implementation_details}, the invariance loss in Appendix~\ref{app:sec:invariance} and give details on data augmentations in Appendix~\ref{app.image_augmentations}. The model pseudo-code is in Appendix~\ref{pseudo_code}. The datasets used in the experiments are freely available from their respective sources.
We also open source SemPPL pretrained checkpoints from our experiments, namely ResNet-50 1×, 2× and 4× as well as ResNet-200 2× together with the evaluation code at  \url{https://github.com/deepmind/semppl}.

{
\small
\bibliography{bibliography}
\bibliographystyle{iclr2023_conference}
}

\appendix
\newpage

\section{Additional results and implementation details}

\label{results_app}

\subsection{Semi-supervised details and results}  \label{app.semi_sup_results}
\paragraph{Implementation details.}
In this work, we follow the protocol of \cite{chen2020big,assran2021semi} for fine-tuning from the first layer of the projector and initialize both the encoder and the first layer of the projector with the parameters of the pretrained model.
We add the randomly initialized classifier on top of the first layer of the projector (after the non-linearity).
We train all the weights (pretrained and classifier weights) using either $1\%$ or $10\%$ of the ImageNet-1k training data, and we use the splits introduced in \cite{chen2020simple} and used in all the methods to compare to \cite{grill2020bootstrap,caron2020unsupervised,dwibedi2021little,lee2021compressive,mitrovic2020representation,relicv2,assran2021semi}.

At training time we randomly crop the image, resize it to $224\times 224$, and then randomly apply a horizontal flip. 
At test time we resize images to $25$6 pixels along the shorter side with bicubic resampling and apply a $224\times 224$ center crop to it. 
Both at training and testing times we subtract from the color channels the average channel value and divide it by the standard deviation of the channel value (as computed on ImageNet-1k). 

We use a cross entropy loss and stochastic gradient descent with Nesterov momentum of $0.9$ to fine-tune the model.
For both $1\%$ and $10\%$ settings, we train for $30$ epochs and decay the initial learning rate by a factor $0.2$ at $18$ and $24$ epochs.
Following the approach of \cite{caron2020unsupervised}, we pick different learning rates for the encoder (and the first projector layer) and for the classifier weights.
We do not use any weight decay or other regularization techniques.
We sweep over batch sizes values in $\{512, 1024, 2048\}$, encoder base learning rate values in $ \{0.005, 0.0035, 0.003, 0.0025, 0.002, 0.001\} $, and linear layer base learning rate values in $\{0.5, 0.3, 0.2, 0.1, 0.05, 0.025\}$.

\begin{table}[ht]
\caption{Top-$1$ and Top-$5$ accuracies (in \%), after semi-supervised fine-tuning with a fraction of ImageNet labels, for a ResNet-50 encoder across a number of representation learning methods.}
\small
\centering
\begin{tabular}{lcccccc}
\toprule
Method & \multicolumn{2}{c}{Top-$1$} & \multicolumn{2}{c}{Top-5} \\
& $1$\% & $10$\% & 1\% & 10\%  \\ \\
\toprule
Supervised~\citep{zhai2019s4l}   & 25.4 & 56.4 & 48.4 & 80.4  \\ 
\midrule
 \emph{Pseudo labels in classification:} \\
 \quad MPL~\citep{pham2021meta}  & - & 73.9 & - & - \\
 \midrule
 \emph{Representation learning methods:} \\
 
 \quad SimCLRv2~\citep{chen2020big}  & 57.9 & 68.4 & - & - \\
 \quad SimCLRv2 + self distillation~\citep{chen2020big} & 60.0 & 70.5 & - & - \\
 \quad CoMatch~\citep{Li2021CoMatchSL} & 66.0 & 73.7 & 86.4 & 91.6 \\
 \quad PAWS~\citep{assran2021semi} & 66.5 & 75.5 & - & - \\
\quad DebiasPL \citep{Wang2022DebiasedLF} & 67.1 & - & 85.8 & - \\
\quad SimMatch \citep{Zheng2022SimMatchSL} & 67.2 & 74.4 & 87.1 & 91.6 \\
\rowcolor{blue!10} \quad \method{} (ours)  & {\bf 68.5} & {\bf 76.0} & {\bf 88.2} & {\bf 92.7} \\
\hline
 \quad SimCLRv2 + Selective Kernels \citep{Li2019SelectiveKN} & 64.5 & 72.1 & 86.7 & 91.4 \\
 \rowcolor{blue!10} \quad \method{} (ours) + Selective Kernels  & {\bf 72.3} & {\bf 78.2} & {\bf 90.6} & {\bf 93.9} \\
\bottomrule
\end{tabular}
\vskip 0.4em
\label{table.semi_sup_r50}
\end{table}

\paragraph{Additional results and larger networks.}

When the architecture of the ResNet-$50$ is modified to include \emph{selective kernels} \citep{Li2019SelectiveKN}, we see significant gains in performance at the expense of additional weights. Our implementation of selective kernels is standard and follows rigorously \citet{Li2019SelectiveKN} for a total of $27.5$ million weights instead of of $25.5$ million for a regular ResNet-$50$. Specifically, we use $2$ channels, two convolution kernels of $(3,3)$ and $(5,5)$ with the latter implemented as a $(3,3)$ dilated convolution with rate $2$, and $32$ grouped convolutions. Unlike SimCLRv2 \citep{chen2020big}, we implement our group convolutions explicitly, and do not use the additional \emph{ResNet-D} architectural modification from \citet{He2019BagOT}.
When using selective kernels our performance after finetuning with $1$\% of labels is the same as that of SimCLRv2 after finetuning with $10$\% of labels.

Additionally, in order to investigate the robustness and scalability of these results, we further test the generality of \method{} by learning representations on larger (both deeper and wider) ResNet encoders.
Table~\ref{table.semi-sup} testifies to \method{} outperforming the competing representation learning methods across all the architectures, both in the $1\%$ and the $10\%$ labelled settings. %
Also, as our flagship result we reach $80.4\%$ top-$1$ accuracy on ResNet-200 $2\times$ with $10\%$ of ImageNet-1k labels. Just as in the ResNet-50 $1\times$ case this figure is comparable with the fully supervised accuracy attained by historical methods. $80.1$\% top-1 is defined as in~\cite{grill2020bootstrap} with standard RandAugment~\citep{Cubuk2020RandaugmentPA} data augmentation. However it's certainly a few percentage accuracy points away from results obtained with optimal current training protocols~\citep{Bello2021RevisitingRI}. We also note that \method{} is pre-trained for $300$ epochs in all cases. This, rather than the $1000$ epochs used as standard by most other representation learning methods, again compares with a typical figure of $200$ epochs used in supervised learning. Overall this hints at \method{} having achieved close to an order of magnitude gain in label efficiency (compared to supervised learning) at a similar epochs budget.

Our final networks were optimized using tranches of between $128$ (for a ResNet-50) and $512$ (for the largest ResNets) Cloud TPUv3s all during $300$ epochs each irrespective of size. This required around a day of computation time per run and tranche for a ResNet-$50$ on $128$ devices, time which scaled approximately linearly with the number of parameters on larger networks, depending on the actual network.

\subsection{Robustness and OOD Generalization}  \label{app.robust_ood}

We test the robustness and out-of-distribution (OOD) generalization abilities of representations learned via \method{} on several detasets. 
We use ImageNetV2~\citep{recht2019imagenet} and ImageNet-C~\citep{hendrycks2019benchmarking} datasets to evaluate robustness and the datasets ObjectNet~\citep{barbu2019objectnet} and
ImageNet-R~\citep{hendrycks2021many} to evaluate the OOD generalization.

The ImageNetV2 dataset~\citep{recht2019imagenet} has three sets of $10000$ images (matched frequency (MF), Threshold 0.7 (T-0.7) and Top Images (TI)) that were collected to have a similar distribution to the ImageNet test set.
The ImageNet-C dataset~\citep{hendrycks2019benchmarking} consists of 15 synthetically generated corruptions of $5$ different severities (e.g. blur, noise) that are applied to the ImageNet validation set. 
The ImageNet-R dataset~\citep{hendrycks2021many} consists of $30000$ different renditions (e.g. paintings, cartoons) of $200$ ImageNet classes; the aim of this dataset is to test the generalization ability to different textures and other naturally occurring style changes that are out-of-distribution to the ImageNet training data. 
The ObjectNet dataset~\citep{barbu2019objectnet} has $18574$ images from differing viewpoints and backgrounds compared to the ImageNet training set.

On all datasets we evaluate the representations learned on a standard ResNet50 encoder under a linear evaluation protocol.
We freeze the pretrained representations (no gradient updates) and train a linear classifier on top of the output of the ResNet-50 encoder using the full labelled ImageNet training set.
We perform the test evaluation zero-shot, i.e the above datasets are not seen during the training of the representation or classifier. 

We provide a detailed breakdown across the different ImageNet-C corruptions in Table \ref{table.imagenet_c_corruptions}. 
Our proposed approach \method{} outperforms both the supervised baseline, on 12 out of 15 corruptions, as well as the competing semi-supervised representation learning model PAWS, on 12 out of 15 corruptions (notably, over all Blur, Weather and Digital corruptions).

\begin{table}[h!]
\caption{Top-1 accuracies (in \%) for OOD generalisation on Gauss, Shot, Impulse, Blur, Weather, and Digital
corruption types of ImageNet-C. \label{table.imagenet_c_corruptions}}
\begin{center}
\begin{adjustbox}{max width=\textwidth}
\begin{tabular}{lccc|cccc|cccc|cccc}
\hline
& & & & \multicolumn{4}{c|}{Blur} & \multicolumn{4}{c|}{Weather} & \multicolumn{4}{c}{Digital}  \\
Method & Gauss & Shot & Impulse  &  Defocus & Glass & Motion & Zoom & Snow & Frost & Fog & Bright & Contrast & Elastic & Pixel & JPEG  \\
\hline
\; Supervised \citep{lim2019fast} & 37.1 & 35.1 & 30.8 & 36.8 & {\bf 25.9} & 34.9 & {\bf 38.1} & 34.5 & 40.7 & 56.9 & 68.1 & 40.6 & {\bf 45.6} & 32.6  & 56.0 \\ 
\hline
\emph{Semi-supervised representations:}   &&& &&&& &&&& &&&&  \\
\; PAWS \citep{assran2021semi} & {\bf 43.5} & {\bf 40.6} & {\bf 33.5} & 38.7 & 19.7 & 34.1 & 32.8 & 40.3 & 44.7 & 64.0 & 70.5 & 59.7 & 42.4 & 38.5 & 55.1 \\
\; \method (ours) & 41.3 & 39.1 & 30.0 & {\bf 41.9} & 23.2 & {\bf 37.5} & 34.0 & {\bf 40.5} & {\bf 45.5} & {\bf 64.4} & {\bf 71.9} & {\bf 60.6} & 44.2 & {\bf 45.1} & {\bf 57.7} \\
\hline
\end{tabular}
\end{adjustbox}
\end{center}
\end{table}

\subsection{Transfer}
\label{sec:supp:transfer}
To further evaluate the usefulness of the learned representations, we evaluate how well they transfer across datasets. For this, we follow the standard evaluation protocol outlined in \cite{grill2020bootstrap,chen2020simple}.
We evaluate \method{} across the linear evaluation protocol which consists of freezing the encoder and only training a randomly initialized linear classifier on top of the encoder. In line with prior work \citep{chen2020simple,grill2020bootstrap,dwibedi2021little}, we test \method{} representations on the following datasets: Food101~\citep{bossard2014food}, CIFAR10~\citep{krizhevsky2009learning}, CIFAR100~\citep{krizhevsky2009learning}, Birdsnap~\citep{berg2014birdsnap}, SUN397 (split 1)~\citep{xiao2010sun}, DTD (split 1)~\citep{cimpoi2014describing}, Cars~\citep{krause20133d} Aircraft~\citep{maji2013fine}, Pets~\citep{parkhi2012cats}, Caltech101~\citep{fei2004learning}, and Flowers~\citep{nilsback2008automated}, where we compare the downstream performance of \method{} to that of other reported semi-supervised methods on 10\% of labels. Across these datasets there are differences in terms of metrics used for selecting the best hyper-parameters as well the reporting of the final results. In line with prior work~\citep{chen2020simple,grill2020bootstrap,dwibedi2021little}, for Food101~\citep{bossard2014food}, CIFAR10~\citep{krizhevsky2009learning}, CIFAR100~\citep{krizhevsky2009learning}, Birdsnap~\citep{berg2014birdsnap}, SUN397 (split 1)~\citep{xiao2010sun}, DTD (split 1)~\citep{cimpoi2014describing}, and Cars~\citep{krause20133d} we report the Top-1 accuracy on the test set, and for Aircraft~\citep{maji2013fine}, Pets~\citep{parkhi2012cats}, Caltech101~\citep{fei2004learning}, and Flowers~\citep{nilsback2008automated} we report the mean per-class accuracy. For DTD and SUN397 we only use the first split of the 10 provided splits in the dataset as per~\cite{chen2020simple,grill2020bootstrap,dwibedi2021little}.

In these experiments, models are initially trained on the training sets of the individual datasets, and the validation sets are used to select the best hyperparameters from the executed hyperparameter sweeps. Once the best hyperparameters have been selected, the final models are trained on a merged dataset containing both the training and the validation split and evaluated on the held-out test split. The final results of the transfer experiments are reported in Table~\ref{table.transfer}. The performed hyperparameter sweeps involved sweeping over the learning rates \{$.001$, $.01$, $0.1$, $0.2$, $0.25$, $0.3$, $0.35$, $0.4$, $1.$, $2.$\}, batch sizes \{$128$, $256$, $512$, $1024$, $2048$\}, weight decay \{$1\mathrm{e}{-6}$, $1\mathrm{e}{-5}$, $1\mathrm{e}{-4}$, $1\mathrm{e}{-3}$, $0.01$, $0.1$\}, warmup epochs \{$0$, $10$\}, momentum \{$0.9$, $0.99$\}, Nesterov \{True, False\}, and the number of training epochs. For the linear transfer protocol we considered setting epochs among \{$20$, $30$, $60$, $80$, $100$\}.  Models were trained by stochastic gradient descent with momentum.

\newpage
\section{Invariance regularization}
\label{app:sec:invariance}

We define the short-hands
\begin{equation}
    p(\hat{z}_m^{a_1};\tilde{z}_{m, t}^{a_2, +} )
    = \frac{\varphi(\hat{z}_m^{a_1}, \tilde{z}_{m, t}^{a_2, +})}{\varphi(\hat{z}_m^{a_1}, \tilde{z}_{m, t}^{a_2, +}) + 
        \sum_{x_n \in\mathcal{N}(x_m)} \varphi(\hat{z}_m^{a_1}, \tilde{z}_{n, t}^{a_2})}
\end{equation}
and 
\begin{equation}
    p(\hat{z}_m^{a_1}; \tilde{z}_{m, t}^{a_2})
    =\frac{\varphi(\hat{z}_m^{a_1}, \tilde{z}_{m, t}^{a_2})}{\varphi(\hat{z}_m^{a_1}, \tilde{z}_{m, t}^{a_2}) + 
        \sum_{x_n \in\mathcal{N}(x_m)}\varphi(\hat{z}_m^{a_1},\tilde{z}_{n, t}^{a_2})}
\end{equation}
where $\mathcal{N}(x_{k})$ is the set of negatives, randomly uniformly sampled from the current batch, $\tilde{z}_{n, t}^{a_2} = g_{t}(f_{t}(x_{n}))$ is the target network projection of the negative sample; $\varphi(x_1, x_2) = \tau \cdot \exp(\langle x_1, x_2 \rangle / \tau)$ is the scoring function, $\tau>0$ is a scalar temperature, and $\langle \cdot, \cdot \rangle$ denotes the standard Euclidean dot product.

We can now rewrite the components of the overall loss  
\begin{equation}
    \mathcal{L}_{\text{\method}} = \mathcal{L}_{\text{\textsc{augm}}}  + \alpha\mathcal{L}_{\text{\textsc{sempos}}}
\end{equation}
as 
\begin{alignat}{1}
\mathcal{L}_{\text{\textsc{sempos}}} = - \sum_{m=1}^{B} 
        \log  p(\hat{z}_m^{a_1};\tilde{z}_{m, t}^{a_2, +} )
\end{alignat}
and
\begin{alignat}{1}
\mathcal{L}_{\text{\textsc{augm}}} = - \sum_{m=1}^{B} 
        \log p(\hat{z}_m^{a_1}; \tilde{z}_{m, t}^{a_2}).
\end{alignat}

As discussed in the main text we add the invariance penalty introduced in \cite{mitrovic2020representation} to further increase the similarity between the anchor and positives and regularize the learning process. 
We add this invariance penalty both for augmentation positives and semantic positives.
In particular, we compute
\begin{align*}
    \mathcal{I}_{augm} = & \kl{p(\hat{z}_m^{a_1}; \tilde{z}_{m, t}^{a_2})}{p(\hat{z}_m^{a_2}; \tilde{z}_{m, t}^{a_1})} \\ =  & sg[\mathbb{E}_{p(\hat{z}_m^{a_1}; \tilde{z}_{m, t}^{a_2})} \log p(\hat{z}_m^{a_1}; \tilde{z}_{m, t}^{a_2})]
    - \mathbb{E}_{p(\hat{z}_m^{a_1}; \tilde{z}_{m, t}^{a_2})} \log p(\hat{z}_m^{a_2}; \tilde{z}_{m, t}^{a_1})
\end{align*}
and 
\begin{align*}
    \mathcal{I}_{sempos} = & \kl{p(\hat{z}_m^{a_1}; \tilde{z}_{m, t}^{a_2, +})}{p(\hat{z}_m^{a_2, +}; \tilde{z}_{m, t}^{a_1})} \\ =  & sg[\mathbb{E}_{p(\hat{z}_m^{a_1}; \tilde{z}_{m, t}^{a_2, +})} \log p(\hat{z}_m^{a_1}; \tilde{z}_{m, t}^{a_2, +})]
    - \mathbb{E}_{p(\hat{z}_m^{a_1}; \tilde{z}_{m, t}^{a_2, +})} \log p(\hat{z}_m^{a_2, +}; \tilde{z}_{m, t}^{a_1})
\end{align*}
where $sg$ denotes the stop-gradient operation. Taking all this together, this gives the final form of the loss as

\begin{equation}
    \mathcal{L}_{\text{\method}} = c(\mathcal{L}_{\text{\textsc{augm}}}  + \alpha\mathcal{L}_{\text{\textsc{sempos}}}) + \lambda(\mathcal{I}_{augm} + \mathcal{I}_{sempos})
\end{equation}
with $\lambda$ the invariance scale and $c$ is the contrastive scale.
We use $\lambda=5$ and $c=0.3$ in all our experiments irrespective of encoder size or training time as our method is robust to the choice of these hyperparameters.

\newpage
\section{Pseudo-code of \method{}}
\label{pseudo_code}

Listing \ref{alg:semppl} provides PyTorch-like pseudo-code for \method{} detailing how we compute pseudo-labels and use them to select the additional semantic positives, which are then used in the contrastive loss, along the augmentation positives.

\noindent\begin{minipage}[t]{0.88\textwidth}
  \begin{algorithm}[H]
\begin{lstlisting}[language=Python, label={alg:semppl}, caption=Pseudo-code for \method{}.]
'''
k: The number of neighbors in k-NN when computing pseudolabels.
f_o: online network: Encoder + comparison_net.
g_t: target network: Encoder + comparison_net.
gamma: Target EMA coefficient.
n_e: Number of negatives.
p_m: Mask apply probability .
'''
for i in range(num_large_views):
    queue_i = queue.init(queue_size, FIFO)

for x, y in batch:  # Load batch of B samples each with data x and (maybe) label y.
    x_m = mask_background(x) 
    for i in range(num_large_views):
        # Stochastically apply background removal.
        x = Bernoulli(p_m) ? x_m : x
        # Create an augmented large view.
        xl_i = augment(crop_large(x)) 
        ol_i = f_o(xl_i)
        tl_i = g_t(xl_i)
        # Enqueue the labeled images in the batch.
        if y is not None:
            queue_i.enqueue((tl_i, y_i))
        
    for i in range(num_small_views):
        xs_i = augment(crop_small(x))
        # Small views only go through the online network
        os_i = f_o(xs_i) 
        
    # Pseudo-label computation for unlabelled examples.
    if y is None:  # Missing label.
        votes = [knn(k, queue_i, ol_j) for i, j in all_pairs(num_large_views)]
        y = mode(votes)
    
    loss = 0  
    # Compute the loss between all the pairs of large views.
    for i in range(num_large_views): 
        for j in range(num_large_views):
            loss += contrastive_loss(ol_i, tl_j, n_e)  # Augmentation positives.
            for _ in range(num_semantic_positives):
                # Sample semantic positives from the queue, and add to the loss.
                z = sample(queue_j.filter(y))
                loss += contrastive_loss(ol_i, z, n_e)
            
    # Compute the loss between the small and large views.
    for i in range(num_small_views):
        for j in range(num_large_views):
            loss += contrastive_loss(os_i, tl_j, n_e)  # Augmentation positives.
            for _ in range(num_semantic_positives):
                # Sample semantic positives from the queue, and add to the loss.
                z = sample(queue_j.filter(y))
                loss += contrastive_loss(ol_i, z, n_e)
                
    loss /= ((num_large_views + num_small_views)
    * num_large_views * (1 + num_semantic_positives))

    # Compute the gradients, and update the online and target network.
    loss.backward()
    update(f_o)
    g_t = gamma * g_t + (1 - gamma) * f_o 
    
\end{lstlisting}
\end{algorithm}
\end{minipage}%
\hfill%

\newpage
\section{Analysis}

\label{analysis_appendix}

\subsection{Implementation details}
\label{additional_analysis_implementation_details}

We perform all the ablation experiments using $10\%$ of labelled data and train a standard ResNet-50 encoder with \method{} for $100$ epochs (except in the training duration ablation). We report the top-1 accuracies on the ImageNet test set after fine-tuning from the first layer of the projector.
As in~\citet{grill2020bootstrap} and for the main results in this paper, we use multi-layer perceptrons for the projector and predictor with $2$ linear layers---the first one followed by batch normalization~\citep{Ioffe2015BatchNA} and rectified linear activation with output sizes $4096$ and $256$ for the two layers respectively. We use the same augmentations as for the experiments in the main paper---the standard \textsc{SimCLR} augmentations~\citep{chen2020simple} and the \alpharelic multi-crop and saliency-based masking~\citep{relicv2}.
Following the hyperparameter settings of the main results, we use 
\begin{itemize}
    \item batch size: $B = 4096$
    \item queue capacity $C=20 B$ (unless specifically ablated)
    \item number of nearest neighbours $k=1$  (unless specifically ablated)
    \item view voting is used (unless specifically ablated)
    \item weight decay: $1e-6$  (exclude biases and batch normalization parameters)
    \item optimizer: LARS (exclude biases and batch normalization parameters from adaptation)
    \item base learning rate: $0.3$ (scaled linearly with batch size~\citep{Goyal2017AccurateLM})
    \item warm-up: $10$ epochs
    \item cosine decay schedule for learning rate
    \item exponential moving average parameter: $0.996$
    \item views: $4$ large views of size $224\times 224$ and $2$ small views of size $96\times 96$
    \item temperature: $\tau=0.2$\
    \item number of semantic positives: $3$ (unless specifically ablated)
    \item $10$ randomly subsampled negatives per anchor
    \item $\alpha=1/5$ (unless specifically ablated), $\lambda=5$ and $c=0.3$.
\end{itemize}

\newpage

\subsection{Additional analyses}
\label{additional_analysis_appendix}

\paragraph{Number of semantic positives}
We study the effect of varying the number of semantic positives in \method{}. Table~\ref{tab.num_pos} shows that increasing this number from $1$ to $3$ only has an effect on the amount of correctly predicted pseudo-labels, but no effect on downstream performance. 
On the other hand, using $5$ or $10$ semantic positives significantly improves performance and also yields much more accurate pseudo-labels prediction.

\paragraph{Training duration}
Next, we vary the length of training representations and examine downstream performance. As can be seen from Table~\ref{tab.length_training}, both the proportion of correctly predicted pseudo-labels and downstream performance improve with longer training up to $300$ epochs but decrease if we continue training up to $500$ epochs. 
This indicates with training longer than $300$ epochs \method{} is starting to overfit, an observation consistent with phenomena reported elsewhere in the literature involving noisy labels~\citep{li2019learning, kim2019nlnl, Liu2020EarlyLearningRP, Han2018CoteachingRT, Zhang2018GeneralizedCE, wang2019symmetric}.

\begin{table}[ht!]
\caption{Top-1 accuracy (in \%) on the ImageNet-1k test set, and accuracy (in \%) of correctly predicted pseudo-labels at the end of training for semantic positives and training length experiments.}
\small
\begin{subtable}[b]{.5\linewidth}
\begin{tabular}{ccc}
\toprule
Num. positives & Top-1 & Pseudo-label acc. \\
\toprule
    $1$  & 69.9 & 68.6 \\
    $2$  & 69.9 & 70.9\\
    \rowcolor{blue!10} $3$  & 69.9 & 69 \\
    $5$  &  {\bf 71.0} & 72.8 \\
    $10$  &  70.7 & {\bf 72.9} \\
\bottomrule
\end{tabular}
\subcaption{Varying the number of semantic positives.}
\label{tab.num_pos}
\end{subtable}%
\begin{subtable}[b]{.5\linewidth}
\begin{tabular}{ccc}
\toprule
Training time (epochs) & Top-1 & Pseudo-label acc. \\
\toprule
    $100$   &  69.9  &  69.2  \\
    $200$   &  72.4  &  76.8  \\
    \rowcolor{blue!10} $300$   &  {\bf 72.7}  &  {\bf 77.9}  \\
    $500$   &  72.3  &  75.6  \\
\bottomrule
\end{tabular}
\subcaption{Varying the length of training.}
\label{tab.length_training}
\end{subtable}
\end{table}

\paragraph{View voting}
\method{} generates multiple views from a single instance image in order to learn representations. Those different views can be leveraged towards better pseudo-labels prediction. Rather than only picking one randomly selected data view to compute a single pseudo-label, we perform \emph{majority voting} over (noisy) pseudo-labels computed from all available image views. Specifically, we compare the online predictor embedding of one view with the queue of the target projector embeddings of the same data view from previous batches in the first setting; in the second setting we compare the online predictor embedding of each view with the queue of the target projector embeddings of each other data view from previous batches. 

Since \method{} relies on $4$ large views, this yields up to $16$ different pairs of views to compare and compute pseudo-labels from, i.e. we get $16$ pseudo-label predictions; this setting we call \emph{view voting}. Table \ref{tab.view_voting} shows that using all available views to compute pseudo-labels significantly increases pseudo-labels accuracy which in turn significantly improves downstream performance.

\begin{table}[ht!]
\caption{Top-1 accuracy (in \%) on the ImageNet-1k test set, and accuracy (in \%) of correctly predicted pseudo-labels at the end of training for view voting experiments (using all views to compute
pseudo-labels vs using just a single view).}
\begin{center}
\small
\begin{tabular}{ccc}
\toprule
View voting & Top-1  & Pseudo-label acc. \\
\toprule
    \rowcolor{blue!10} On & \textbf{69.9} & \textbf{68.6} \\
    Off & 69.0 & 62.5 \\
\bottomrule
\end{tabular}
\label{tab.view_voting}
\end{center}
\end{table}

\paragraph{Number of nearest neighbours}
In order to compute pseudo-labels we use $k$-nearest neighbour lookup on the queue. While in the main results Section 3 we consistently assume $k=1$ here we ablate the effect of varying $k$ on downstream performance. As can be seen Table \ref{tab.knn_number}, increasing $k$ actually leads to a small \emph{decrease} in performance. This is attributable to the decrease in the proportion of correctly predicted pseudo-labels as $k$ increases.

\paragraph{Queue length}
How long should the queue of target projector embeddings for computing pseudo-labels be? As the queue grows in size, it contains increasingly stale embeddings and threatens to hamper the accuracy of predicted pseudo-labels. On the other hand, increasing queue size increases the amount and diversity of labelled embeddings available which we would expect to be beneficial.
Those two opposing forces---staleness and increasing coverage---govern the accuracy with which we can correctly predict pseudo-labels in turn directly affecting the selection of semantic positives and their quality. 
We resolve this ambiguity empirically.
As seen in Table \ref{tab.queue_length}, increasing the coverage and diversity of labelled embeddings has a strong positive effect on representation learning and downstream performance. Staleness of embeddings is far less of a problem at practical (i.e., not very large) queue sizes, showing diversity and coverage to be the dominant factor.

\begin{table}[ht!]
\caption{Top-1 accuracy (in \%) on the ImageNet-1k test set, and accuracy (in \%) of correctly predicted pseudo-labels at the end of training for $k$-NN and queue length experiments.}
\small
\begin{subtable}[b]{.48\linewidth}
\centering
\begin{tabular}{ccc}
\toprule
\textbf{k}-nn & Top-1 & Pseudo-label acc. \\
\midrule
    \rowcolor{blue!10} $1$ & \textbf{70.0} & \textbf{70.5} \\
    $2$ & 69.9 & 69.5 \\
    $3$ & 69.8 & 70.0\\
    $5$ & 69.8 & 70.0 \\
    $10$ & 69.1 & 68.0\\
\bottomrule
\end{tabular}
\subcaption{Varying the number of nearest neighbours.}
\label{tab.knn_number}
\end{subtable}%
\hfill
\begin{subtable}[b]{.48\linewidth}
\centering
\begin{tabular}{ccc}
\toprule
Queue size (C) & Top-1 & Pseudo-label acc.  \\
\midrule
    $4000$ & 67.9 & 53.1 \\
    $8000$ & 68.6 & 60.4 \\
    $12000$ & 68.8 & 62.0\\
    $20000$ & 69.2 & 64.3\\
    $40000$ & 69.6 & 66.9 \\
    \rowcolor{blue!10} $80000$ & \textbf{69.9} & 68.8 \\
    $200000$ & 69.8 & \textbf{69.5}\\
\bottomrule
\end{tabular}
\subcaption{Varying queue size.}
\label{tab.queue_length}
\end{subtable}
\end{table}

\paragraph{The effect of the loss weight $\alpha$}

We use $\alpha$ in Equation~\ref{eq:final_loss_sempos} to weight the contribution of the loss coming from semantic positives against the loss coming from augmentation positives. In \method{} we use multiple semantic positives for learning and thus we need to adjust $\alpha$ to appropriately weigh the contribution of the individual semantic positives. In our main experiments, we use $3$ semantic positives and for this reason $\alpha$ is not equal to $1$.

In Table~\ref{tab.alpha} we vary $\alpha$ from $0.0$ to $1.0$, with $\alpha=0.0$ effectively recovering \alpharelic. The results indicate that the \method{} loss notably improves over \alpharelic, but that the exact choice of $\alpha$ should be treated as a hyperparameter. Note that the value $0.2$ is very close to the $\frac{1}{3}$ which is exactly $1 / \textrm{number of semantic positives}$. Thus, the optimal choice of $\alpha$ is very close to that fraction.

\begin{table}[ht!]
\caption{Top-1 accuracy (in \%) on the ImageNet-1k test set for varied values of the loss weight $\alpha$}
\begin{center}
\small
\begin{tabular}{lc}
\toprule
Loss weight ($\alpha$) & Top-1 \\
\toprule
    $0.0$ (\alpharelic) & 72.4 \\
    \rowcolor{blue!10} $0.2$ & \textbf{76.0} \\
    $1.0$ & 74.6 \\
\bottomrule
\end{tabular}
\label{tab.alpha}
\end{center}
\end{table}

\newpage

\section{Augmentations}  \label{app.image_augmentations}

In this work, we follow the established data augmentations protocols and pipelines of \cite{chen2020simple, grill2020bootstrap, caron2020unsupervised,mitrovic2020representation,chen2020big}.
Specifically, \method{} uses a set of augmentations to generate different views of the original image which has three channels, red $r$, green $g$ and blue $b$ with $r,g,b \in [0,1]$.
 
The augmentations we use are generated by applying the following sequence of operations in the following order
\begin{enumerate}
    \item Crop the image: Randomly select a patch of the image, between a minimum and maximum crop area of the image, with aspect ratio sampled log-uniformly in $[3/4, 4/3]$. Upscale the patch, via bicubic interpolation, to a square image of size $s \times s$.
    \item Flip the image horizontally.
    \item Colour jitter: randomly adjust brightness, contrast, saturation and hue of the image, in a random order, uniformly by a value in $[-a, a]$ where $a$ is the maximum adjustment (specified below).
    \item Grayscale the image, such that the  channels are combined into one channel with value $0.2989r + 0.5870g + 0.1140b$.
    \item Randomly blur. Apply a $23\times 23$ Gaussian kernel with standard deviation sampled uniformly in $[0.1, 2.0]$.
    \item Randomly solarize: threshold each channel value such that all values less than $0.5$ are replaced by $0$ and all values above or equal to $0.5$ are replaced with $1$.
\end{enumerate}
Apart from the initial step of image cropping, each subsequent step is applied with a certain probability to generate the augmented view of the original image. 
These probabilities and other augmentation parameters are given in Table~\ref{table:aug}. 
\method{} uses $4$ large views of size $224\times 224$ pixels and $2$ small views of $96\times 96$ pixels; to get the first and third large views and the first small view we use the parameters listed below for odd views, while for the second and fourth large view and the second small view we use the parameters for even views.
Note that these are the same augmentations used also in \cite{chen2020simple, grill2020bootstrap, caron2020unsupervised,mitrovic2020representation,chen2020big}.

In addition to these augmentations, we also randomly apply the saliency masking augmentation proposed in \cite{relicv2} which enables us to remove a large part of the background. 
We follow the protocol described in \cite{relicv2} for computing the saliency masking for an image and we apply this augmentation with probability $0.1$ to the $4$ large views.
In keeping with \cite{relicv2}, we fill out the removed background of the image with homogeneous grayscale noise with the grayscale level randomly sampled for each view.
We only apply the saliency masking when the remaining foreground covers at least $20\%$ of the total image.

\begin{table}[ht]
\begin{center}
    \begin{tabular}{lcc}
    \midrule
    \textbf{Parameter} & Even views & Odd views \\
    \midrule
    Probability of randomly cropping & 50\% & 50\% \\
    Probability of horizontal flip & 50\% & 50\% \\
    Probability of colour jittering & 80\% & 80\% \\
    Probability of grayscaling & 20\% & 20\% \\
    Probability of blurring & 100\% & 10\% \\
    Probability of solarization & 0\% & 20\% \\
    Maximum adjustment $a$ of brightness & 0.4 & 0.4 \\
    Maximum adjustment $a$ of contrast & 0.4 & 0.4 \\
    Maximum adjustment $a$ of saturation & 0.2 & 0.2 \\
    Maximum adjustment $a$ of hue & 0.1 & 0.1 \\
    Crop size $s$ & 224 & 96 (small), 224 (large) \\
    Crop minimum area & 8\% & 5\% (small), 14\% (large) \\
    Crop maximum area & 100\% & 14\% (small), 100\% (large) \\
    \bottomrule
    \end{tabular}
\end{center}
\caption{Parameters of data augmentation scheme. Small/large indicates small or large crop.}
\label{table:aug}
\end{table}

\newpage
\section{Computational cost of \method{}}
\label{ccost_app}

\paragraph{Added cost of \method{}}

As \method{} is based on \alpharelic{}, the computational overhead of \method{} over \alpharelic{} comes from three factors: i) the queue maintenance, ii) the $k$-NN execution, and iii) the computation of the added loss term.
We carefully analyzed this overhead, and concluded the following:

\begin{itemize}
    \item The $k$-NN and the queue maintenance take $5.6\%$ of the step time, with the $k$-NN itself taking $4.5\%$.
    \item The total loss (both forward and backward passes of loss terms coming from both augmentation and semantic positives) takes $3.9\%$ of the step time, with $2.9\%$ belonging to the additional loss coming from semantic positives.
    \item In total, \method{} takes 8.5\% of the total step time.
\end{itemize}

Note that in our work we use a naive implementation of the $k$-NN. If needed, the $k$-NN can be further sped up, at a fraction of the accuracy, with a fast approximate $k$-NN model such as FAISS~\citep{johnson2019billion}.
This would further reduce the computation cost of using the $k$-NN to predict pseudo-labels and select semantic positives.
Additionally, since our code is not heavily optimised, there is room for additionally lowering these numbers with targeted optimisations.

\paragraph{Comparing \method{} to related models}

The direct comparison of training times of \method{} to related models can be deceiving, since there are three different sources of difference that influence the comparison. These summarise to differences in: i) accelerator architectures used, ii) deep learning frameworks used, and iii) concrete code, which can be assumed, is not optimised for any of the models, given their research nature. Being aware of these sources of difference, we contrast the running times of pre-training for \method{} and related models:

\begin{itemize}
    \item PAWS~\citep{assran2021semi} report $8.5$ hours per $100$ epochs on $64$ V100s. Assuming continuity in speed, this is equivalent to $25.5$ hours for $300$ epochs.
    \item SimMatch~\citep{Zheng2022SimMatchSL} report $2.34$ hours per epoch on $8$ V100s. If we (generously) assume perfect scaling (which is difficult in reality) to $64$ V100s, this is equivalent to $87.75$ hrs for $300$ epochs.
    \item \method{} trains in $13$ hours on $64$ TPUv3 cores.
\end{itemize}

Given empirical evidence that TPUv3 is roughly $23\%$ faster than V100 on ResNet-50 training~\citep{tpu_vs_gpu_etc}, one could infer that the implementation of our model is significantly more computationally efficient. However, given that we cannot control all the aforementioned differences mentioned above, more careful analysis is needed to establish the right cost for each of these methods.

\end{document}